\documentclass[runningheads]{llncs}
\usepackage[T1]{fontenc}
\usepackage{graphicx}
\usepackage{booktabs}
\usepackage{multirow}
\usepackage{amsmath}
\usepackage{url}
\usepackage{xcolor}
\usepackage{pifont}
\usepackage{tabularx}
\usepackage{siunitx}

\begin{document}
\title{GAIDA: Generative AI for Human-in-the-Loop Document Annotation}
\titlerunning{GAIDA}
\author{Islam Mesabah\inst{1,2} \orcidID{0000-0002-2298-2127} \and Ali Elhosary\inst{2} \orcidID{0009-0007-5694-5694} \and Priyabanta Sandulu\inst{1,2} \orcidID{0009-0003-9284-5093} \\Darko Obradovic\inst{1,2} \orcidID{0009-0002-2681-5649} \\ \and Sebastian J. Vollmer\inst{1,2} \orcidID{0000-0003-2831-1401}}
\authorrunning{I. Mesabah et al.}
\institute{
German Research Center for Artificial Intelligence (DFKI),\\
\and
RPTU Kaiserslautern-Landau,\\
Kaiserslautern, Germany\\
\email{islam.mesabah@dfki.de}
}
\maketitle

\begin{abstract}
Document annotation is a fundamental yet labor-intensive step in building document information extraction systems. While Optical Character Recognition (OCR) and Large Language Models (LLMs) can automate parts of this process, we argue that annotation systems should be judged not by how completely they remove human effort, but by how effectively they redirect it toward review and correction. We propose a modular human-in-the-loop architecture that integrates OCR-based extraction, LLM-based semantic pre-annotation, and human validation in a single editable workflow, and present \textbf{GAIDA}, a research prototype serving as its reference implementation. GAIDA supports interchangeable OCR engines, LLM backends, and prompt configurations, and enables stage-specific measurement of annotation effort. In an initial single-annotator, stage-specific study on the FUNSD scanned-form and WildReceipt retail-receipt benchmarks, OCR assistance reduced region-creation and transcription time by up to \textbf{80.9\%}, and reviewing LLM pre-annotations reduced semantic-labeling time by up to \textbf{73.7\%}, despite imperfect pre-annotations of 78.86\% and 86.98\% Micro F1 on FUNSD and WildReceipt, respectively. Finally, we reuse the same implementation, adapting only OCR and prompt configuration, to build and openly release \textbf{AraReceipt}, a manually annotated dataset of 100 Arabic receipt images labeled with 25 key-information classes. The results indicate that imperfect automated outputs provide practical value when presented as editable suggestions under human control.

\end{abstract}

\section{Introduction}
Document information extraction transforms semi-structured documents into structured representations for downstream systems. Building high-quality annotated datasets remains a bottleneck~\cite{yu2023documentnet,zhou2022radically}: identifying regions, transcribing content, assigning labels, and validating output is slow and prone to inconsistency under complex layouts~\cite{ma2024mmlongbench,liao2025minority}.

OCR and LLMs can restructure, not merely accelerate, this process: OCR localizes and transcribes text automatically, after which an LLM proposes semantic labels for human validation, shifting the annotator's role from creation to verification~\cite{schroeder-etal-2025-just,zhang2025augmenting}. Model accuracy alone does not determine practical value; usefulness also depends on the correction effort created by OCR errors, prompt design, model choice, and the review interface. Studying these factors requires a common modular architecture in which individual components can be exchanged while the surrounding annotation workflow and human-review process remain fixed.

To operationalize this architecture, we developed \textbf{GAIDA}\footnote{\url{https://github.com/islammesabah/gaida}}, a research prototype that serves as its reference implementation. GAIDA integrates OCR-based extraction, LLM-based semantic pre-annotation, and human validation within the open-source Label Studio\footnote{\url{https://labelstud.io/}} annotation environment. GAIDA rests on the premise that annotation systems should be judged by how effectively they redirect human effort toward review, not by how completely they remove it. We evaluate GAIDA in two complementary document settings: FUNSD~\cite{jaume2019funsddatasetformunderstanding} represents noisy forms whose labels often depend on relations between regions, whereas WildReceipt~\cite{sun2021spatialdualmodalitygraphreasoning} represents dense retail receipts with a substantially more fine-grained field schema. We also reuse the same architecture to construct and publicly release \textbf{AraReceipt}\footnote{\url{https://huggingface.co/datasets/IslamMesabah/AraReceipt}}, a dataset of 100 Arabic receipt images derived from ReceiptSense~\cite{abdallah2024receiptsense} and annotated using a 25-class schema. Accordingly, we position this work as a system-and-resource contribution accompanied by an initial single-annotator feasibility study; a controlled multi-annotator evaluation is left to future work.

This paper addresses four research questions:
\begin{itemize}
    \item \textbf{RQ1:} To what extent does OCR assistance reduce region-creation and transcription time, and how much localization/recognition error remains for human correction?
    \item \textbf{RQ2:} How sensitive is semantic pre-annotation quality to model choice and prompt design under a fixed workflow?
    \item \textbf{RQ3:} Does reviewing LLM-generated pre-annotations reduce semantic-labeling time relative to manual assignment?
    \item \textbf{RQ4:} Which components and configurations must be adapted to apply GAIDA to Arabic receipts?
\end{itemize}

\textbf{Contributions:} (1) a modular human-in-the-loop annotation architecture, together with \textbf{GAIDA}, a research prototype that serves as its reference implementation and supports interchangeable OCR and LLM components; (2) initial stage-specific evidence that OCR assistance reduces region-creation and transcription time, while LLM pre-annotations reduce semantic-labeling time; (3) a sensitivity analysis showing that model choice affects pre-annotation quality far more than prompt elaboration, and that more prescriptive prompting does not consistently help (Sect.~\ref{subsec:rq2}); and (4) \textbf{AraReceipt}, a manually annotated, publicly released Arabic dataset of 100 images and 25 classes, showing architectural reuse with language-specific configuration only.

\section{Related Work}
Document information extraction benchmarks rely on task-specific annotations
that may include document regions, transcriptions, semantic entities, and
key--value relations~\cite{naparstek2024kvp10k}. Creating such resources
requires clear guidelines, annotator training, agreement assessment, and
quality control~\cite{klie2024annotation,ding2026deep}, since inconsistent
annotation schemes can undermine benchmark
reliability~\cite{laatiri2023redundancy}. OCR can pre-extract text regions and
transcriptions~\cite{saout2024overview,yoon2024lofi}; however, recognition and
localization errors, as well as choices in representing OCR output, can
propagate to downstream
processing~\cite{vanstrien2020assessing,colakoglu2025problem,shen2026ocr}.
This motivates evaluating not only standalone OCR accuracy but also the
correction effort left to human annotators.

LLMs increasingly generate structured predictions from OCR output or document
images~\cite{colakoglu2025problem,shen2026ocr}. Recent work incorporates
document layout in different ways: DocLLM models textual and spatial
information jointly and LayIE-LLM studies configuration choices for
layout-rich information-extraction
pipelines~\cite{wang2024docllm,colakoglu2025problem}. These studies show that
extraction quality depends on input representation, prompt design, annotation
schema, and model
choice~\cite{colakoglu2025problem,shen2026ocr}.
Model-centered performance scores, however, do not quantify the human effort
required to inspect and correct the resulting outputs.

Human-in-the-loop annotation therefore treats machine-generated pre-labels as
editable suggestions rather than final labels, retaining human control over
correction and approval~\cite{wang2024humanllm,goel2023llms}. Such pre-labels
can reduce expert annotation time~\cite{goel2023llms,wang2024humanllm},
although incorrect pre-annotations can bias annotator decisions or shift
effort from label creation to error
detection~\cite{wang2024humanllm,pangakis2025keeping}. Closest to our setting,
DocSpiral integrates document processing, annotation interfaces, and iterative
model improvement in a human-in-the-spiral workflow, progressively reducing
manual effort through repeated retraining~\cite{sun2025docspiral}. GAIDA is
complementary: it holds the assisting models fixed and instead decomposes
\emph{where} human effort is spent---region creation, transcription
correction, and semantic labeling---under interchangeable OCR and LLM
components. To our knowledge, this stage-specific decomposition of annotation
effort has not been reported for OCR--LLM-assisted document annotation.

Multilingual document understanding introduces further challenges from script
variation, uneven OCR performance across languages, and limited annotated
resources~\cite{yoon2024lofi,yang2024ccocr,heakl2025kitab}. Multilingual
systems and resources include LOFI, a language-, OCR-, and form-independent
extraction pipeline evaluated on Korean and Japanese business
documents~\cite{yoon2024lofi}, and AMuRD and ReceiptSense, which provide
Arabic--English receipt
data~\cite{abdallah2023amurd,abdallah2024receiptsense}. These works motivate a
complementary system-level question: whether the same human-in-the-loop
implementation can be reused across scripts through configuration changes
rather than architectural redesign.

GAIDA is thus positioned as a research prototype and reference implementation
of a modular OCR--LLM--human annotation architecture: rather than a new
extraction model or a model-training platform, it provides a common editable
workflow for comparing component and prompt configurations, measuring
transcription and semantic-labeling effort separately, and examining
configuration-level reuse for Arabic receipts.

\section{Proposed Architecture and GAIDA Implementation}
\label{sec:methodology}

The proposed architecture follows a human-in-the-loop design in which automated components assist, but do not replace, the human annotator. OCR and LLM inference produce editable pre-annotations that a human inspects, corrects, and approves before export. GAIDA implements this architecture as a modular research prototype, separating OCR processing, semantic labeling, human validation, and export into interchangeable components.

\subsection{Workflow and Architecture}
\label{subsec:workflow}

\begin{figure}[t]
    \centering
    \includegraphics[width=0.9\linewidth]{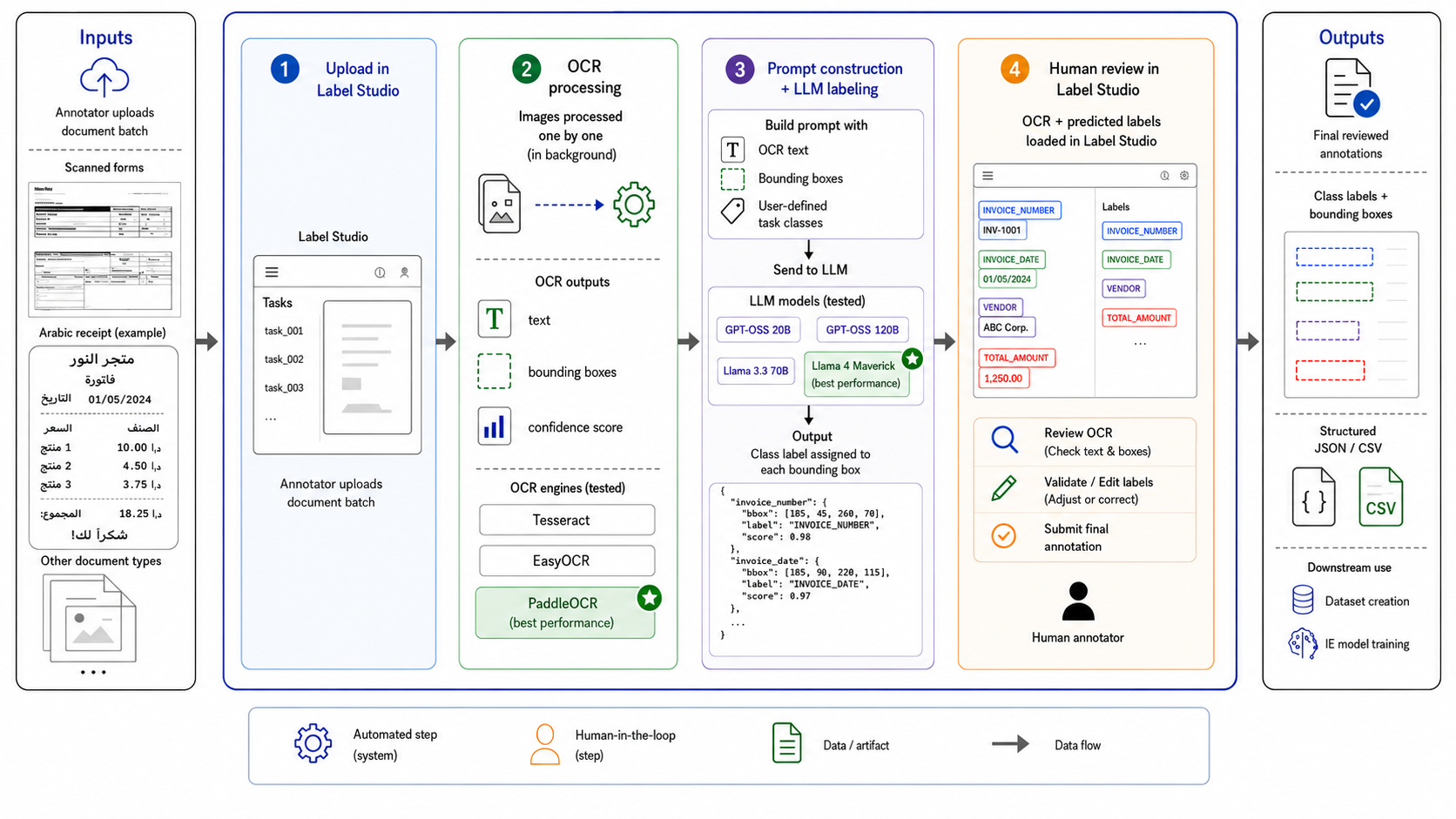}
    \caption{The proposed human-in-the-loop annotation workflow as implemented in GAIDA. OCR extracts text regions and transcriptions, an LLM generates preliminary semantic labels, and a human annotator reviews, corrects, and approves each prediction before export.}
    \label{fig:gaida-overview}
\end{figure}

As shown in Fig.~\ref{fig:gaida-overview}, the workflow implemented in GAIDA follows four stages. First, the annotator defines the valid semantic classes and annotation configuration in Label Studio. Second, the selected OCR engine returns text regions, transcriptions, and confidence values. Third, GAIDA constructs a structured prompt containing the recognized text, corresponding text regions, valid label set, and task instructions; the LLM then assigns one semantic class to each region. Finally, the resulting pre-annotation is loaded into Label Studio, where the annotator reviews and corrects the regions, transcriptions, and labels before export. Label Studio provides the annotation interface, while the GAIDA backend coordinates OCR processing, prompt construction, LLM inference, response parsing, and transfer of the resulting annotations. OCR engines need only expose text, coordinates, and confidence; LLM backends need only accept a structured prompt and return machine-readable labels. This lets the same architecture be reused, unmodified, for AraReceipt (Sect.~\ref{subsec:rq4}).

\subsection{OCR Processing and Normalization}
\label{subsec:ocr-processing}

Each OCR engine returns a set of regions $\mathcal{R}=\{(b_i,t_i,c_i)\}_{i=1}^{n}$, where $b_i$, $t_i$, and $c_i$ denote the bounding box, transcription, and confidence value of region $i$, respectively. GAIDA converts engine-specific bounding-box formats into a common image-relative coordinate representation, removes leading and trailing whitespace from recognized text, and discards detections with missing text or invalid coordinates. This normalization allows OCR backends to be exchanged without modifying the downstream prompt-construction or annotation components. It also prevents differences in engine-specific output formats from introducing spurious variation, while preserving genuine differences arising from the engines' detection and recognition behavior and from document properties such as layout complexity, image quality, and region density (Sect.~\ref{subsec:rq1}). Recognition errors require textual correction, whereas localization errors---including missing, duplicated, split, merged, or incorrectly bounded regions---require structural correction; the two error types are therefore reported separately (Sect.~\ref{subsec:rq1}).

\subsection{Prompt Construction and Development}
\label{subsec:prompting}

GAIDA provides the LLM with the complete OCR output, including each text region $t_i$ and its bounding box $b_i$, together with the valid label set $\mathcal{Y}$ and task instructions $P$. The model returns a machine-readable mapping that assigns one label $\hat{y}_i \in \mathcal{Y}$ to each OCR region. We first compared multiple LLMs on FUNSD using the same \textbf{rule-based prompt}. This prompt defined the valid classes, provided class-specific rules and tie-breaking criteria, and required the model to assign exactly one label to each OCR region. Based on this comparison, we selected the best-performing model for the subsequent prompt experiments.

Using the selected model, we then evaluated three prompt families on FUNSD and WildReceipt. The \textbf{label-definition prompt} clarified the meaning and scope of each class to improve distinctions between semantically similar labels. The \textbf{decision-order prompt} further extended these definitions by introducing a priority sequence for cases in which multiple labels appeared plausible. Finally, the \textbf{dataset-refined prompt} was developed through an exploratory qualitative review of preliminary predictions. Recurring misclassifications and ambiguous label distinctions were identified from the model outputs and translated into additional dataset-specific rules and clarifications. Because this process used errors observed on the evaluated data rather than a held-out development split, the resulting scores represent iterative prompt engineering on these datasets and should not be interpreted as evidence of out-of-sample generalization.

The complete pre-annotation is finally presented for human review: the annotator verifies regions, corrects boundaries and transcriptions, and confirms or changes each label; export happens only after this review, letting us test whether correction is faster than annotating from scratch.

\section{Experimental Evaluation}
\label{sec:evaluation}

We combine model-centered measurements of OCR and semantic-labeling quality with annotator-centered measurements of elapsed time. The timing experiments are preliminary stage-specific comparisons: OCR and semantic labeling were timed independently, rather than as one continuous end-to-end annotation session. RQ1--RQ3 use within-annotator comparisons on FUNSD and WildReceipt; RQ4 uses the construction of AraReceipt (Sect.~\ref{subsec:rq4}).

\subsection{Experimental Design}
\label{subsec:experimental-design}

We use two English benchmarks (Table~\ref{tab:datasets}). FUNSD~\cite{jaume2019funsddatasetformunderstanding} contains noisy scanned forms with a compact four-class schema (\textit{header}, \textit{question}, \textit{answer}, \textit{other}), where class assignment often depends on relations between neighboring regions. WildReceipt~\cite{sun2021spatialdualmodalitygraphreasoning} contains dense retail receipts with a 25-class schema, where classes are more often cued by local lexical or positional signals than by inter-region relations. This distinction is interpreted in Sect.~\ref{sec:discussion}.

\begin{table}[t]
\centering
\caption{Datasets used in the study. Test image counts refer to benchmark test
sets; for timing, we used 50 FUNSD and 20 WildReceipt images.}
\label{tab:datasets}
\small\begin{tabularx}{\textwidth}{p{2.3cm}lccX}
\toprule
\textbf{Dataset} & \textbf{Document Type} & \textbf{Test Images} &
\textbf{Classes} & \textbf{Role in the Study} \\
\midrule
FUNSD~\cite{jaume2019funsddatasetformunderstanding} & Scanned forms & 50 & 4 & OCR evaluation, model selection, prompt development, and timing. \\
\addlinespace
WildReceipt~\cite{sun2021spatialdualmodalitygraphreasoning} & Retail receipts & 472 & 25 & OCR, prompt, and timing evaluation under a dense layout and fine-grained schema. \\
\addlinespace
AraReceipt (Ours) & Arabic receipts & 100 & 25 & Cross-language application and public Arabic dataset construction. \\
\bottomrule
\end{tabularx}
\end{table}

To isolate each assistance stage, we compare four workflows (Table~\ref{tab:workflows}), pairing a manual and an assisted condition for each pipeline stage. In every condition, the annotator worked toward the visible benchmark ground truth and was timed until producing a matching annotation; final quality is thus matched by construction, and the reported times isolate effort rather than quality. The semantic-labeling comparison specifically began both conditions from the same ground-truth OCR regions and transcriptions, isolating the LLM's contribution from OCR quality.

\begin{table}[h]
\centering
\caption{Comparison of the annotation workflows evaluated in this study.}
\label{tab:workflows}
\small\begin{tabular}{p{2.4cm}c@{\hspace{2mm}}c@{\hspace{2mm}}p{3.5cm}@{\hspace{2mm}}p{4.3cm}}
\toprule
\textbf{Workflow} & \textbf{OCR} & \textbf{LLM} & \textbf{System Output} & \textbf{Primary Human Activity} \\
\midrule
manual-OCR & \ding{55} & \ding{55} & None & Create regions/text. \\
\addlinespace
assisted-OCR & \ding{51} & \ding{55} & Regions, transcriptions & Validate, correct regions/text. \\
\addlinespace
manual-labeling & \ding{55} & \ding{55} & None & Create labels. \\
\addlinespace
assisted-labeling & \ding{55} & \ding{51} & Labels & Validate, correct labels. \\
\bottomrule
\end{tabular}
\end{table}

All sessions followed the same fixed and non-randomized order: for each dataset, a single annotator completed manual OCR, assisted OCR, manual labeling, and assisted labeling on the same images, with a one-week gap between the manual and assisted passes. Using the same annotator removed between-annotator variability, while the one-week interval helped reduce immediate memory and fatigue effects. Because the assisted condition always followed the manual condition on the same images, familiarity effects are possible (Sect.~\ref{sec:discussion}).

WildReceipt timing used a random 20-image subset of its 472 test images because of the study's time constraints, whereas FUNSD timing used all 50 test images. The same selected images were used in the corresponding manual and assisted conditions. Timing was conducted by one annotator, a native Arabic speaker with professional English proficiency. The same annotator completed all English timing sessions and annotated AraReceipt. This controls between-annotator speed variation but does not remove the possible influence of non-native English proficiency on the absolute English timing baselines (Sect.~\ref{sec:discussion}).

We integrated three OCR engines into GAIDA---Tesseract~\cite{1288165.1288167}, EasyOCR\footnote{\url{https://github.com/JaidedAI/EasyOCR}}, and PaddleOCR~\cite{cui2025paddleocr}---to verify that the architecture could support different OCR backends without changing the surrounding workflow. Following a preliminary qualitative inspection of their outputs, PaddleOCR was selected for the quantitative experiments (Table~\ref{tab:ocr-results}). Because this screening was not a formal comparative benchmark, we do not report or claim quantitative superiority over Tesseract or EasyOCR. For semantic pre-annotation, we evaluated GPT-OSS 20B, GPT-OSS 120B~\cite{agarwal2025gpt}, Llama~3.3~70B~\cite{grattafiori2024llama}, and Llama~4 Maverick\footnote{\url{https://ai.meta.com/blog/llama-4-multimodal-intelligence/}}. All models were hosted locally and served through Ollama using a temperature of 0.0, a maximum output length of 2,048 tokens, and a timeout of 180 seconds. Because temperature was fixed at 0.0, each model--prompt configuration was evaluated once; the reported results therefore do not estimate run-to-run variation. Model comparison was conducted exclusively on FUNSD (Table~\ref{tab:model-comparison}). Llama~4 Maverick was selected for the prompt experiments because it achieved the highest Macro F1 under the shared rule-based prompt. For the semantic-labeling timing experiment, the assisted condition used pre-annotations generated by Llama~4 Maverick with the dataset-refined prompt for the corresponding dataset.

\subsection{Evaluation Metrics}
\label{subsec:metrics}

We evaluate OCR localization using intersection over union, $\operatorname{IoU}(\hat{b},b)=|\hat{b}\cap b|/|\hat{b}\cup b|$, between predicted and ground-truth bounding boxes $\hat{b}$ and $b$. Detection is scored by greedy one-to-one matching at $\tau_d=0.5$: all pairs with $\operatorname{IoU}\geq0.5$ are ranked by descending IoU and accepted sequentially if neither box is already matched, so a predicted region must overlap a ground-truth region substantially---while tolerating minor boundary differences---and no prediction receives credit for more than one ground-truth region. Given the accepted matches $M$ and the ground-truth and predicted region sets $G$ and $P$, recall is $|M|/|G|$ and precision is $|M|/|P|$; Mean IoU, $|M|^{-1}\sum_{(g,p)\in M}\operatorname{IoU}(g,p)$, measures boundary quality of detected regions rather than coverage.

Segmentation consistency uses a relaxed threshold of 0.1, which identifies weak overlaps associated with possible splitting or merging and is not used to determine correct detections. With $c_g=\sum_p\mathbf{1}[\operatorname{IoU}(g,p)\geq0.1]$ and $c_p=\sum_g\mathbf{1}[\operatorname{IoU}(g,p)\geq0.1]$ counting the predictions overlapping ground-truth region $g$ and the ground-truth regions overlapping prediction $p$, $\operatorname{SplitRate}=\sum_g c_g/|G|$ and $\operatorname{MergeRate}=\sum_p c_p/|P|$ give the mean overlap counts per ground-truth region and per prediction; both are unbounded above, and values below 1 indicate unmatched regions. We also report $\operatorname{Exact1to1}=\sum_g\mathbf{1}[c_g=1]/|G|$, the proportion of ground-truth regions overlapping exactly one prediction at this threshold, without enforcing bidirectional one-to-one matching. Transcription quality is evaluated by the Character Error Rate, $\mathrm{CER}=(S+D+I)/N$, with $S$, $D$, and $I$ the character substitutions, deletions, and insertions and $N$ the reference length, averaged over matched regions per image and then macro-averaged over images.

Semantic pre-annotation is evaluated on the ground-truth OCR regions and transcriptions using Micro and Macro F1; because each ground-truth region receives a predicted label, these scores measure label agreement independently of OCR localization and transcription errors. Micro F1 aggregates true positives, false positives, and false negatives across labels; Macro F1 is $\frac{1}{|\mathcal{Y}|}\sum_{y\in\mathcal{Y}}\mathrm{F1}_y$, the unweighted mean of per-label F1. Annotation time is the elapsed time per image to produce an annotation consistent with the ground truth (Sect.~\ref{subsec:experimental-design}), and time reduction is $(T_{\mathrm{manual}}-T_{\mathrm{assisted}})/T_{\mathrm{manual}}\times100$.

\subsection{RQ1: OCR-Assisted Annotation}
\label{subsec:rq1}

Table~\ref{tab:ocr-results} summarizes PaddleOCR performance across the evaluated datasets. WildReceipt achieves higher detection precision and recall, together with a lower CER, indicating better localization and transcription than FUNSD. FUNSD's higher SplitRate (1.18 vs.\ 0.96) is consistent with more gold regions having multiple loosely-overlapping candidates; WildReceipt's higher MergeRate (1.08 vs.\ 1.02) is consistent with adjacent items collapsing into single detections.

\begin{table}[h]
\centering
\caption{PaddleOCR results. Detection metrics use one-to-one matching at $\operatorname{IoU}\geq0.5$, while overlap metrics use $\operatorname{IoU}\geq0.1$ (Sect.~\ref{subsec:metrics}). Arrows indicate the preferred direction.}
\label{tab:ocr-results}
\small
\begin{tabular}{l@{\hspace{2mm}}lc@{\hspace{2mm}}c}
\toprule
\textbf{Task Type} & \textbf{Metric} & \textbf{FUNSD} &
\textbf{WildReceipt} \\
\midrule
\multirow{3}{*}{Detection and localization}
 & Precision $\uparrow$ & 65.00\% & 83.88\% \\
 & Recall $\uparrow$ & 73.00\% & 75.80\% \\
 & Mean IoU (matched) $\uparrow$ & 75.00\% & 76.50\% \\
\midrule
\multirow{3}{*}{Overlap structure}
 & Exact 1-to-1 Ratio $\uparrow$ & 83.00\% & 87.97\% \\
 & Split Rate $\rightarrow 1$ & 1.18 & 0.96 \\
 & Merge Rate $\rightarrow 1$ & 1.02 & 1.08 \\
\midrule
Transcription & CER $\downarrow$ & 8.60\% & 6.70\% \\
\bottomrule
\end{tabular}
\end{table}

Recognition errors require textual correction, whereas localization and overlapping errors require regions to be resized, added, removed, split, or merged. As shown in the upper part of Table~\ref{tab:combined-time}, the assisted-OCR condition substantially reduces this correction effort compared with manual-OCR: from 13:30 to 3:37 for FUNSD (73.2\%) and from 8:49 to 1:41 for WildReceipt (80.9\%). The longer manual annotation time observed for FUNSD is consistent with its lower detection precision and greater split tendency; however, the effects of region count, text length, layout structure, and session order (Sect.~\ref{subsec:experimental-design}) were not disentangled.

\begin{table}[h]
\centering
\caption{Manual vs.\ assisted annotation time per image, by pipeline stage
(mm:ss), for the workflows defined in Table~\ref{tab:workflows}. Times measure
correction toward visible benchmark ground truth (Sect.~\ref{subsec:experimental-design});
$n$ = timed images per dataset.}
\label{tab:combined-time}
\small
\begin{tabular}{llccc}
\toprule
\textbf{Stage} & \textbf{Dataset} & \textbf{Manual} & \textbf{Assisted} &
\textbf{Reduction} \\
\midrule
\multirow{2}{*}{OCR / transcription}
 & FUNSD ($n=50$)       & 13:30 & 3:37 & \textbf{73.2\%} \\
 & WildReceipt ($n=20$) & 8:49  & 1:41 & \textbf{80.9\%} \\
\midrule
\multirow{2}{*}{Semantic labeling}
 & FUNSD ($n=50$)       & 1:53  & 0:36 & \textbf{68.1\%} \\
 & WildReceipt ($n=20$) & 1:39  & 0:26 & \textbf{73.7\%} \\
\bottomrule
\end{tabular}
\end{table}

\textbf{RQ1:} OCR assistance reduced region-creation and transcription time by 73.2\% and 80.9\% on FUNSD and WildReceipt, respectively, although localization and recognition errors still required manual correction (Sect.~\ref{sec:discussion} discusses ordering effects).

\subsection{RQ2: Sensitivity to Model and Prompt Configuration}
\label{subsec:rq2}

We first compare the LLMs on FUNSD using a shared \textbf{rule-based} prompt (Table~\ref{tab:model-comparison}). Performance varies widely: Micro F1 ranges from 40.57\% (GPT-OSS~20B) to 72.43\%, which Llama~3.3~70B and Llama~4~Maverick both reach. Maverick attains the higher Macro F1 (63.77\% vs.\ 62.26\%), indicating more balanced performance across classes, and is therefore selected for the prompt experiments.

\begin{table}[h]
\centering
\caption{Semantic pre-annotation performance across models on FUNSD using the
same rule-based prompt.}
\label{tab:model-comparison}
\small\begin{tabular}{lcc}
\toprule
\textbf{Model} & \textbf{Micro F1 (\%)} & \textbf{Macro F1 (\%)} \\
\midrule
GPT-OSS~20B      & 40.57 & 39.15 \\
GPT-OSS~120B     & 67.20 & 56.38 \\
Llama~3.3~70B    & \textbf{72.43} & 62.26 \\
Llama~4 Maverick & \textbf{72.43} & \textbf{63.77} \\
\bottomrule
\end{tabular}
\end{table}

Using Maverick, we next vary the prompt (Table~\ref{tab:prompt-engineering}). On FUNSD, explicit label definitions raise Micro F1 to 74.23\%, whereas the decision-order prompt \emph{lowers} it to 70.80\% (62.45\% Macro F1); the dataset-refined prompt is strongest at 78.86\% Micro and 70.18\% Macro F1. On WildReceipt, whose 25-class schema is far more fine-grained than FUNSD's four classes and whose prompts therefore include more detailed rationales, the three prompt families are nearly indistinguishable: Micro F1 varies by less than 0.6 points (86.42--86.98\%) and Macro F1 by less than 0.3 points (77.15--77.41\%), with no configuration dominating on both metrics.

\begin{table}[h]
\centering
\caption{Semantic pre-annotation performance across prompt configurations using
Llama~4 Maverick.}
\label{tab:prompt-engineering}
\small\begin{tabularx}{\columnwidth}{l X cc}
\toprule
\textbf{Dataset} & \textbf{Prompt Configuration} &
\textbf{Micro F1 (\%)} & \textbf{Macro F1 (\%)} \\
\midrule
\multirow{3}{*}{FUNSD}
 & Label-definition prompt & 74.23 & 64.15 \\
 & Decision-order prompt   & 70.80 & 62.45 \\
 & Dataset-refined prompt  & \textbf{78.86} & \textbf{70.18} \\
\midrule
\multirow{3}{*}{WildReceipt}
 & Label-definition prompt & 86.42 & 77.15 \\
 & Decision-order prompt   & 86.60 & \textbf{77.41} \\
 & Dataset-refined prompt  & \textbf{86.98} & 77.36 \\
\bottomrule
\end{tabularx}
\end{table}

\textbf{RQ2:} Pre-annotation quality is far more sensitive to model choice than to prompt elaboration. Under an identical prompt, Micro F1 on FUNSD spans 40.57--72.43\% across models---a gap no prompt variant approaches. Prompt design is inconsistent by comparison: label definitions help, a fixed decision order hurts on FUNSD, and all prompt families converge on WildReceipt. More prescriptive prompting is therefore not inherently better; only guidance derived from observed errors (the dataset-refined prompt) consistently helped. Because that prompt was refined on the evaluated data itself, its scores reflect iterative engineering rather than out-of-sample generalization.

\subsection{RQ3: Human-in-the-Loop Efficiency}
\label{subsec:rq3}

The lower part of Table~\ref{tab:combined-time} reports the initial stage-specific comparison of manual-labeling and assisted-labeling under the configuration specified in Sect.~\ref{subsec:experimental-design}. Both conditions began from the same ground-truth OCR regions and transcriptions, thereby isolating the contribution of semantic pre-annotation. On FUNSD, the average time per image decreased from 1:53 to 0:36, corresponding to a 68.1\% reduction. On WildReceipt, it decreased from 1:39 to 0:26, corresponding to a 73.7\% reduction.

Because the annotator corrected each image toward visible ground truth (Sect.~\ref{subsec:experimental-design}), these reductions quantify the effort to reach a known target rather than independent annotation quality. \textbf{RQ3:} Reviewing LLM pre-annotations reduced semantic-labeling time by 68.1\% and 73.7\%, despite pre-annotation Micro F1 of only 78.86\% and 86.98\%---imperfect suggestions sufficed to redirect effort from creation to correction (threats to validity in Sect.~\ref{sec:discussion}).

\subsection{RQ4: Cross-Language Workflow Adaptation}
\label{subsec:rq4}
To examine cross-language applicability, we constructed \textbf{AraReceipt} from 100 Arabic receipt images selected from ReceiptSense~\cite{abdallah2024receiptsense}. A native Arabic speaker annotated the images without independent review, so we describe the dataset as manually annotated rather than independently validated (Sect.~\ref{sec:discussion}).

The core architecture---annotation interface, normalized data representation,
OCR--LLM communication, human-validation workflow, and export---was reused
without modification, and the PaddleOCR and Llama~4~Maverick configurations
were retained without language-specific fine-tuning. Adaptation was limited to
configuration: OCR settings for Arabic and mixed-script text, and prompt
revisions for Arabic content, right-to-left writing direction, and
language-specific field expressions.

Table~\ref{tab:arareceipt-stats} summarizes the resulting dataset: 100 images
with 4{,}609 annotated regions (46.1 per image), labeled with 25
key-information classes plus an \textsc{Ignore} category, following the
WildReceipt schema. Slightly over half of the regions contain Arabic script
and 43.6\% are purely numeric (prices, quantities, totals), reflecting the
mixed-script character of Arabic retail receipts; key-type classes are
markedly rarer than their value counterparts, as Arabic receipts often print
values without field labels. AraReceipt is released under the MIT license,
matching the terms of ReceiptSense, from which its images are derived.

End-to-end GAIDA-assisted annotation, including OCR review, semantic-label
correction, and final approval, averaged 5:51 per image ($\approx$10 hours in
total); since no manual Arabic baseline was timed, this is an assisted cost
only. \textbf{RQ4:} GAIDA transferred to Arabic receipts through
configuration-level adaptation alone, with no changes to the architecture or
implementation logic, demonstrating cross-language reusability, though not
efficiency gains over a manual Arabic baseline.

\begin{table}[t]
\centering
\caption{AraReceipt statistics. Class counts show the five most and least
frequent labels; full per-class counts accompany the dataset release.}
\label{tab:arareceipt-stats}
\small\begin{tabular}{lr@{\hspace{1.8em}}lr}
\toprule
\textbf{Property} & \textbf{Value} & \textbf{Class} & \textbf{Regions} \\
\midrule
Images                & 100            & \textsc{Others}              & 1{,}598 \\
Annotated regions     & 4{,}609        & \textsc{Prod\_price\_value}  & 466 \\
Regions/image         & 46.1 $\pm$ 21.6 & \textsc{Prod\_quantity\_value} & 461 \\
Arabic-script regions & 52.0\%         & \textsc{Prod\_item\_value}   & 402 \\
Numeric-only regions  & 43.6\%         & \textsc{Subtotal\_value}     & 225 \\
Latin-script regions  & 4.0\%          & \ldots                       &     \\
                      &                & \textsc{Tax\_key}            & 21  \\
                      &                & \textsc{Tips\_key}           & 10  \\
                      &                & \textsc{Tips\_value}         & 9   \\
                      &                & \textsc{Store\_name\_key}    & 1   \\
\bottomrule
\end{tabular}
\end{table}

\section{Discussion}
\label{sec:discussion}

Across the four questions, a consistent picture emerges. Pre-annotation quality depends on OCR reliability, model choice, and---to a lesser and less consistent degree---prompting strategy (RQ1, RQ2). Reviewing imperfect pre-annotations nonetheless substantially reduced stage-specific annotation time (RQ3), supporting the premise that value comes from redirecting human effort from creation to targeted correction rather than from eliminating it. RQ4 shows the same workflow transfers to Arabic receipts through configuration alone.

The two benchmarks also suggest where each assistance stage pays off. FUNSD, with sparse relational structure and noisy scans, incurred the higher manual region-creation cost (13:30 vs.\ 8:49 per image) and gained most in absolute terms from OCR assistance, while its four relational classes kept manual labeling comparatively slow despite the small schema. WildReceipt, with dense but locally-cued fields, was cheaper to transcribe manually yet showed the larger relative gains from both stages. This is consistent with a simple heuristic---layout-noisy, relational documents benefit most from OCR pre-extraction, whereas dense, lexically-cued documents benefit most from label pre-annotation---though our two-dataset, single-annotator design can motivate this hypothesis, not establish it.

\textbf{Threats to validity and future work.}
\label{subsec:threats-validity}
\begin{itemize}\itemsep2pt
\item \emph{Annotator and ordering.} The timing study involved one bilingual
annotator in a fixed manual-before-assisted order. The one-week gap between
passes and identical target annotations in both conditions mitigate immediate
memory and quality-difference effects, but residual familiarity may inflate
the assisted-condition gains. Multiple annotators and counterbalanced session
orders would separate assistance effects from familiarity and speed
variability.
\item \emph{Stage-specific timing.} OCR and semantic labeling were timed
independently, so the reported reductions characterize each assistance stage
but cannot be combined into an end-to-end workflow estimate. Larger studies
should time continuous sessions, cover the full WildReceipt test set, add an
AraReceipt manual baseline, and report variance statistics from the per-image
logs.
\item \emph{Evaluation-set tuning.} The dataset-refined prompts were developed
from errors observed on the evaluated FUNSD and WildReceipt sets; their scores
therefore reflect iterative engineering rather than out-of-sample performance.
The model comparison and the label-definition and decision-order prompts were
fixed before evaluation and are unaffected. Held-out development splits would
enable generalization claims.
\item \emph{Dataset validation.} AraReceipt was annotated by one native Arabic
speaker without independent review. A second annotation pass and
inter-annotator agreement analysis would strengthen its validation.
\end{itemize}
Beyond these remedies, future work will extend the evaluation across languages, OCR engines, and LLM backends, and analyze interaction logs to identify which OCR and labeling errors contribute most to correction effort.

\section{Conclusion}
\label{sec:conclusion}
We proposed a modular human-in-the-loop annotation architecture and introduced GAIDA as its research-prototype reference implementation, separating OCR processing, LLM-based semantic pre-annotation, and human validation into interchangeable components with stage-specific effort measurement. In initial timing experiments on FUNSD and WildReceipt, OCR assistance reduced region-creation and transcription time by up to 80.9\%, and reviewing LLM-generated labels reduced semantic-labeling time by up to 73.7\%, despite pre-annotations reaching only 78.86\% and 86.98\% Micro F1, respectively. Sensitivity analysis showed model choice mattered far more than prompt elaboration: more prescriptive prompts did not consistently help, and only guidance derived from observed errors did. These results reflect an initial single-annotator, fixed-order, stage-specific study in which labeling was timed against visible ground truth and the dataset-refined prompts were tuned on the evaluated sets (Sect.~\ref{sec:discussion}). We further applied GAIDA's unmodified architecture, adapting only OCR and prompt configuration for script and layout, to build and publicly release AraReceipt, a 100-image, 25-class Arabic receipt dataset. Within these bounds, the results support the central premise of the architecture: pre-annotations need not be accurate to be valuable when they redirect human effort from creation to targeted correction.

\bibliographystyle{splncs04}
\bibliography{mybibliography}

\end{document}